# Bio-Skin: A Cost-Effective Thermostatic Tactile Sensor with Multi-Modal Force and Temperature Detection

Haoran Guo[1], Haoyang Wang[1], Zhengxiong Li[2], *Member, IEEE,* Lingfeng Tao[3], *Member, IEEE*

*Abstract*— **Tactile sensors can significantly enhance the perception of humanoid robotics systems by providing contact information that facilitates human-like interactions. However, existing commercial tactile sensors focus on improving the resolution and sensitivity of single-modal detection with high-cost components and densely integrated design, incurring complex manufacturing processes and unaffordable prices. In this work, we present Bio-Skin, a cost-effective multi-modal tactile sensor that utilizes single-axis Hall-effect sensors for planar normal force measurement and bar-shape piezo resistors for 2D shear force measurement. A thermistor coupling with a heating wire is integrated into a silicone body to achieve temperature sensation and thermostatic function analogous to human skin. We also present a cross-reference framework to validate the two modalities of the force sensing signal, improving the sensing fidelity in a complex electromagnetic environment. Bio-Skin has a multi-layer design, and each layer is manufactured sequentially and subsequently integrated, thereby offering a fast production pathway. After calibration, Bio-Skin demonstrates performance metrics—including signal-to-range ratio, sampling rate, and measurement range—comparable to current commercial products, with one-tenth of the cost. The sensor's real-world performance is evaluated using an Allegro hand in object grasping tasks, while its temperature regulation functionality was assessed in a material detection task.**

## I. Introduction

Tactile sensors play a crucial role in humanoid robot systems by providing physical contact information that enables adaptive interaction and manipulation. These sensors allow robots to safely and effectively perform complex tasks, such as disaster relief, hazardous material handling, and telesurgery. As humanoid robots are expected to become increasingly widespread, their large-scale production and adoption bring critical requirements related to cost and functionality. Specifically: (1) for large-scale deployment, it is essential to minimize unit costs while ensuring efficient manufacturability and fabrication; and (2) since humanoid robots will directly interact with humans, their tactile sensors should not only detect force but also sense temperature and regulate its temperature similar to the human skin.

Current commercial tactile sensors for humanoid robots feature complex designs incorporating expensive components, leading to manufacturing challenges and high costs, typically exceeding $10,000. Like the BioTac [1] sensor employs a conductive liquid that is difficult to encapsulate and seal, resulting in increased production line costs; similarly, the uSkin [2][3] sensor uses multiple three-axis Hall-effect sensors, with raw material costs exceeding those of single-axis versions by more than tenfold. Furthermore, Meta's latest DIGIT sensor [4], developed for high-precision detection with optical sensors, has significantly higher unit costs due to the high-cost optical sensors and integrated processing. However, a bio-mechanical study [5] demonstrates that the human finger's tactile resolution and sensitivity are much lower than some current tactile sensors while still supporting human's dexterous manipulation stills. Furthermore, according to the Engineering Cybernetics Theory [6] of Tsien, system design should adopt simple and cost-effective solutions that will always bring more robust performance under the condition of meeting performance requirements. The Raptor engine of SpaceX is a good example of adopting this theory. Compared to the first generation, the third generation Raptor achieves a 1.5 thrust-to-weight ratio increase with a much simpler structure at only one-tenth of the cost. This principle of achieving superior performance through simplified, cost-effective design is not limited to aerospace; it can be equally transformative in developing tactile sensors.

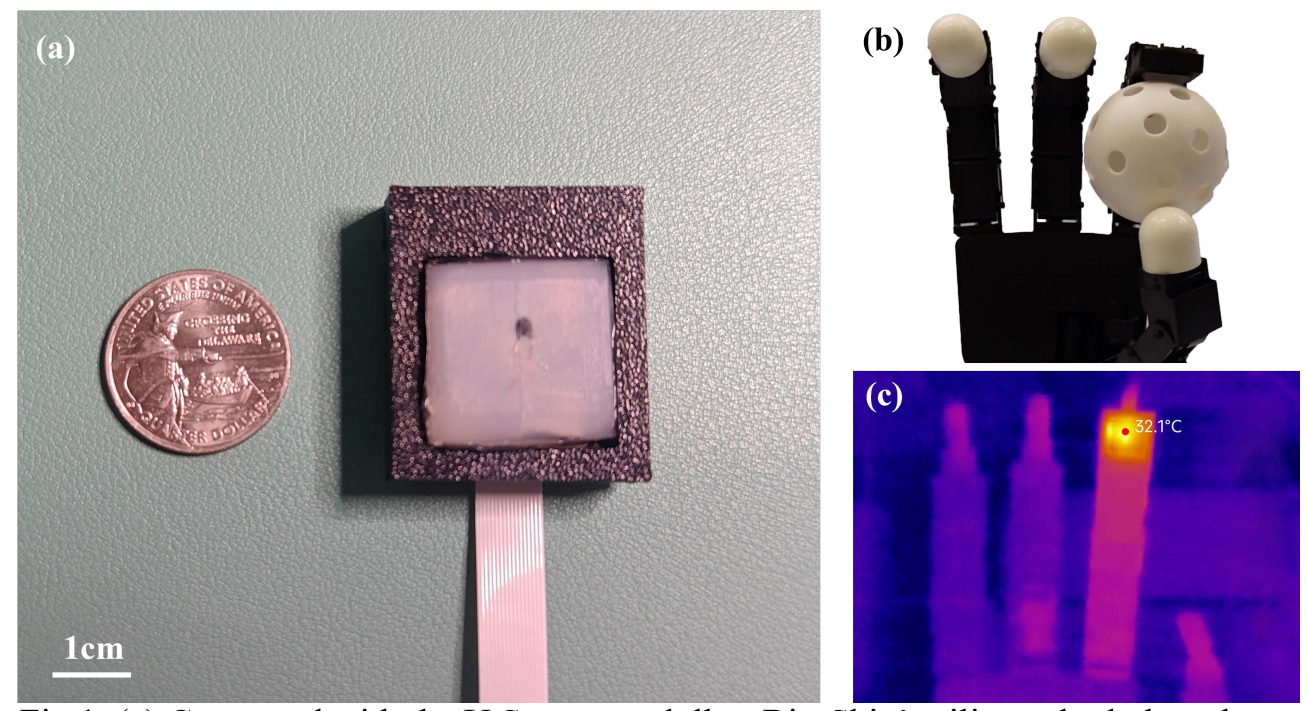

Fig.1. (a) Compared with the U.S. quarter dollar, Bio-Skin's silicone body has almost same size with the coin. (b) Bio-Skin can be directly installed on commercial robot hand such as Allegro hand and Leap hand. (c) Bio-Skin maintaining human skin temperature captured by thermal camera.

Besides, recent tactile sensor research has primarily concentrated on single-modal sensing with the same type of sensor, such as hall effect, piezoresistors, capacitance, and optical sensors. However, these designs are highly susceptible to environmental interference. For instance, uSkin [2][3] and some other designs [7][8] exclusively employ Hall-effect sensors, which are prone to disruption in strong magnetic fields. Also, Digit [4] optical sensors measure the deformation of the external shell, which has a narrow measurement range, while easily losing all signals when the field of view is occupied. In contrast, multi-modal sensors [9] based on different sensing principles possess excellent anti-interference capabilities, making self-validation through multi-modal outputs possible to ensure data reliability. However, due to the cost, most of these designs are limited to lab environments.

In this work, we present Bio-Skin, a cost-effective, multi-modal thermostatic tactile sensor (Fig.1), which integrates five

This work is supported by US NSF grants #2426469 and #2426470.
[1]H. Guo and H. Wang are with Oklahoma State University, Mechatronics and Intelligent Robotics Lab, 563 Engineering North, Stillwater, OK 74075 (e-mail: haoran.guo; haoyang.wang@okstate.edu)
[2]Z. Li is with University of Colorado Denver, Department of Computer Science and Engineering, 1380 Lawrence St. Center, LW-834, Denver. CO 80217 (e-mail: zhengxiong.li@ucdenver.edu)
[3]L. Tao is with the Department of Robotics and Mechatronics Engineering, Kennesaw State University, 840 Polytechnic Lane, Marietta, GA 30060 (e-mail: ltao2@kennesaw.edu)

single-axis Hall-effect sensors for planar normal force sensing and four piezoresistors for 2D shear force sensing. Temperature sensing plays a critical role in detecting extreme surface and environmental temperatures that may pose a risk to the robot. Recognizing the importance of temperature sensing in achieving multi-modal capabilities akin to human skin, we incorporate an additional sensing modality by embedding a thermistor coupled with a heating wire, which enables both temperature sensing and regulation, allowing robots to accurately perceive object materials in the thermal domain and adapt their interactions accordingly. Bio-Skin has a multi-layer design, and each layer is manufactured sequentially and subsequently integrated, offering a fast fabrication pathway. It can be easily integrated with robotic hand systems such as the Allegro Hand. Furthermore, the multiple sensing principles bring a cross-reference framework that enhances force sensing fidelity by validating coupled normal and shear force signals using Hall-effect and piezoresistive sensing mechanisms, which improves sensor robustness in complex electromagnetic environments. We also add data processing and visualization tools for real-time analysis to enhance usability. Once accepted, the sensor design and software package will be open sourced on GitHub to facilitate further development and adoption. We validate the Bio-Skin on the Allegro Hand, demonstrating its effectiveness in object grasping and temperature retention tasks, achieving performance comparable to existing commercial products.

The contributions of this work are summarized as follows:

1). Compared to existing tactile sensors that exceed $10,000, Bio-Skin is significantly more cost-effective, costing approximately one-tenth of these sensors, paving the path for mass production and adoption.

2). Designed Bio-Skin with multi-modal sensing, including normal, shear force, and temperature sensing. A heating function is being integrated into a tactile sensor for the first time, enabling thermostatic control and enhancing realistic interaction akin to human skin.

3). Fabricate the sensor prototype and design the evaluation metrics and calibration process.

4). Develop a cross-reference method to ensure data fidelity by validating the force sensing with two sensing modalities.

5). Evaluate the sensor prototype in object grasping on an Allegro hand and material detection tasks

## II. Related Work

### A. Commercial Tactile Sensor

Robotic hands in commercial applications are typically outfitted with tactile sensors that enable complex task execution, yet they often suffer from high costs and are limited in single-modal function. The uSkin [2][3] sensor utilizes a three-axis Hall sensor array to measure normal and shear forces. This reliance on costly and single-modal sensor arrays limits both market viability and anti-interference ability. In another approach, Meta's Digit [4] and other researchers' work [10] rely on optical technology to detect sensor shell deformations through light distortion, providing superior resolution in forces sensing. However, they face inevitable cost problems due to the usage of optical sensors. This optical sensor also brings a complex system structure, making it hard to use in tiny places. BioTac [1] is a pioneering commercial tactile sensor that combines rigid metal cores with conductive fluids to provide multi-modal feedback, particularly in vibration sensing. However, BioTac's design relies on conductive fluids, which will cause complex manufacturing requirements and high costs.

### B. Multi-modal Tactile Sensor in Literature

Most of the current tactile sensors are single-modal designs, such as Hall-effect-based, vision-based tactile sensors [11][12][13]. These sensors have low reliability in complex electromagnetic environments. Tactile sensors with multi-modal functions can provide high-fidelity feedback to real-world Robots in complex environments, enabling robust and stable interactions. P. Weiner et al. introduced a multi-modal sensor system [14]for robotic hands that utilizes the Hall effect and TOF sensors to measure the force, distance, joint angle, acceleration, and temperature. P. Mittendorfer introduced a self-organizing, multi-modal artificial skin system [9][15] which integrates force, proximity, temperature, and acceleration sensors. However, current multi-modal tactile sensors focus on improving performance instead of balancing performance and cost. Such an objective usually causes complex integration and bulky sensor size that, in turn, limits the sensor installation and the number of sensors. In our approach, we design the Bio-Skin based on engineering cybernetics theory [6] to find the best sensing modality for the tactile sensor and balance performance and cost.

## III. Bio-Skin Sensor Design and Fabrication

### A. Sensor Profile

The sensor profile is essential in guiding the design of Bio-Skin. In this work, we define the following three key principles based on the functionality of human skin:

**1). Multi-modal Sensing Capability** – The Bio-Skin should detect multiple stimuli, including normal force, shear force, and temperature, to enhance sensory feedback.

**2). Thermoregulation** – It should maintain a temperature similar to the real human skin to improve user comfort.

**3). Soft Body and Adaptive Contact Surface** – A compliant body and contact surface should be incorporated to ensure seamless interaction with various objects and surfaces.

Additionally, to promote widespread adoption, the Bio-Skin design should emphasize:

- **Cost-Effective Components** – Using affordable materials to minimize sensor production costs.
- **Simple Fabrication Process** – Enabling fast production and reducing overall manufacturing expenses.

We will utilize cost-effective materials and components suitable for mass production to meet the outlined requirements, incorporating 3D printing and PCB customization to optimize fabrication. Bio-Skin is designed to replicate the sensory capabilities of human fingertips, targeting a normal force sensing range of approximately 1 - 6 N, shear force detection between 1–10 N, and temperature sensing from -10 to 40°C while maintaining a self-regulated temperature of around 30 - 34°C. This approach ensures a balance between performance, scalability, and affordability, making Bio-Skin a viable solution for widespread adoption.

### B. Sensor Modality

**Normal force:** Hall-effect sensors are widely used for force detections as they can capture subtle magnetic field variations, enabling high-precision force measurement.

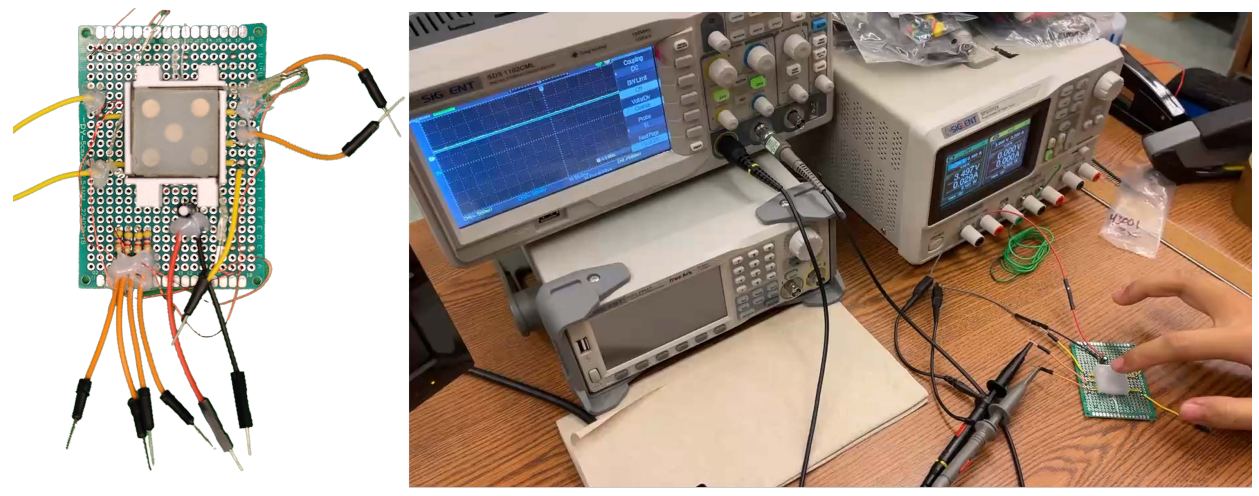

Fig.2 Proof-of-concept of the tactile sensor system with Hall-effect sensors, piezoresistors tested using an oscilloscope for force signal.

TABLE I. COMPONENTS LIST OF BIO-SKIN

| Hall-effect sensor | Controller | Magnet | Labor Cost |
|---|---|---|---|
| DRV5056A4QDBZR ($1.5/each sensor) | ESP32-S3 ($2.4/each sensor) | Neodymium5x2mm ($1/each sensor) | 4~10h ($60~150/each sensor) |
| **MUX** | **Silicone Gel** | **PCB** | **Thermistor** |
| CD74HC4067M96 ($0.35/each sensor) | Dragon Skin 10 Fast ($0.5/each sensor) | JLC PCB ($2/each sensor) | 100kΩ ($0.2/each sensor) |

There are 3-axis Hall-effect sensors that can simultaneously detect normal and shear forces by capturing 3D magnetic field changes. However, their cost is approximately ten times that of 1-axis sensors, increasing material costs significantly if using multiple sensors. In this work, the Bio-Skin sensor utilizes five 1-axis Hall-effect sensors to detect the normal planar forces on the sensor surface, with five magnets associated with the sensors to provide the magnetic field.

**Shear force:** film piezo resistors are chosen due to their cost-effectiveness and suitability for large-scale production. Unlike Hall-effect sensors, piezoresistors measure a material's change in electrical resistance when mechanical stress is applied. When shear force is applied, the material deforms, causing a change in resistance. This change is related to the applied force, allowing for sensing shear force. A different modality eliminates mutual interference and achieves noise detection while enabling a cross-reference function. Meanwhile, their simple structure reduces fabrication complexity and offers high design flexibility.

**Temperature sensing and regulation:** a thermistor is integrated into Bio-Skin to monitor temperature. Meanwhile, a heating wire is inserted to achieve the thermoregulation function. These cost-effective sensors and components benefit from well-established large-scale production technology, making them a practical choice for Bio-Skin.

### C. Proof of Concept

We first developed a proof-of-concept system to validate the proposed sensor modality. Five Hall-effect sensors were mounted on a universal PCB, with one at the center and four at the corners to optimize contact surface coverage. These sensors were housed within a 3D-printed frame, while four film piezo resistors were affixed to their inner surfaces for shear force detection. A silicone body embedded with five magnets in the bottom layer was aligned with the Hall sensors. Force signals are obtained from the piezoresistors via a voltage divider and directly from the Hall sensors. The assembled sensor was tested using an oscilloscope, as shown in Fig.2, while impulse forces were applied to the silicone body. The results exhibited smooth voltage transitions and distinct responses from all nine outputs, confirming the feasibility of the selected components for force sensing.

The thermistor and heating wire are tested with an adjustable power supply and a thermal camera. The results show the thermistor can read the accurate temperature in 5 s, and the heating wire can heat up the sensor body to maintain a temperature range of around 30 - 34°C with 2 w of power.

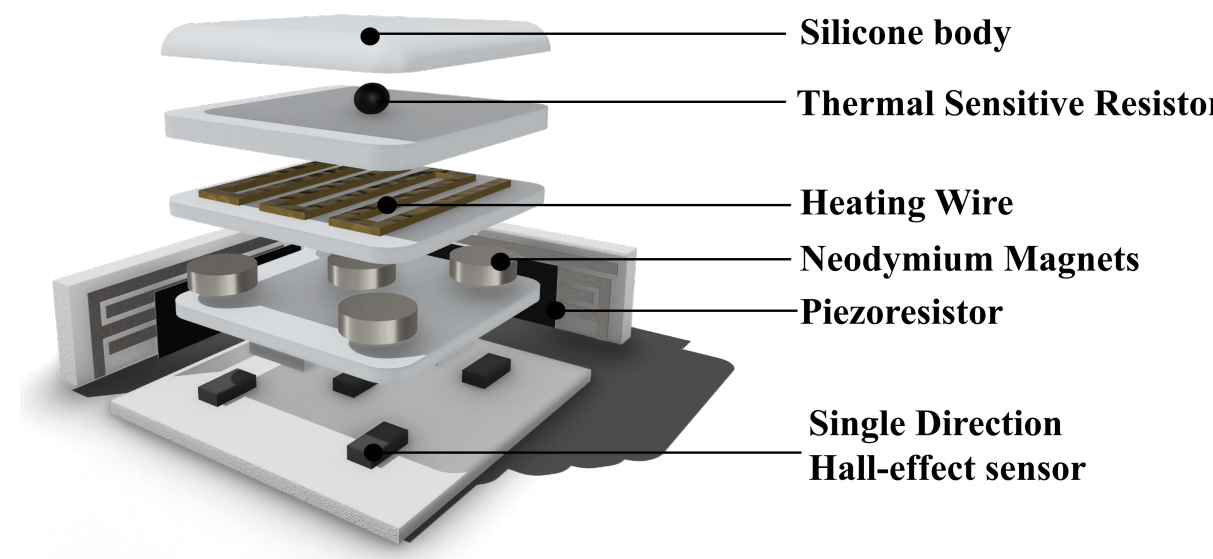

Fig.3 Exploded view of Bio-Skin sensor, showcasing normal force, shear force, and temperature regulation layers with Hall-effect sensors, piezoresistors, magnets, heating and sensing components.

To select the most appropriate components, we tested Hall-effect sensors, piezoresistor from different brands, and various heating wires with different resistances. The final components and cost of Bio-skin are listed in Table I. The cost for each sensor is under 10 USD without labor, and all material not listed in the table are all lower than 0.1 USD.

### D. Bio-Skin Sensor Package Design

Bio-Skin's design can be divided into three layers: the normal force layer, the shear force layer, and the temperature regulation layer. Details are shown in Fig.3. It should be noted that the current design is to fit the size of Allegro hand, but this multi-layer design can be easily scaled up to different sizes and resolutions according to requirements.

**Normal force layer (bottom):** This layer features a custom 26×26 mm PCB as the Bio-Skin base, with soldering positions for five Hall-effect sensors and reserved spots for a piezoresistor, temperature control, and other components. Four 3 mm diameter holes are located at the PCB corners for secure frame attachment. Above the PCB, a silicone body embeds five magnets in an identical distribution. Four low-profile springs are positioned beneath the silicone body to ensure a consistent gap between the magnets and the PCB.

**Shear force layer (Surround):** This layer features a fixing frame that restricts the silicone body's movement while integrating four film piezo resistors. The frame has internal dimensions of 20×20×12 mm with 4 mm thickness — reinforced to 6 mm on one side for mounting on the Allegro Hand. Each corner features a 3 mm diameter, 5 mm long cylindrical extension for precise alignment with the PCB's fixing holes. Four securely attached film piezoresistors (20×12 mm) inside the frame ensure stable sensor placement.

**Temperature regulation layer (Top):** This layer consists of a heating wire and a thermistor embedded within the silicone body. The thermistor is positioned on the upper surface for temperature sensing, while the heating wire is placed beneath the thermistor for better heat spreading.

### E. Fabrication

The optimized fabrication processing is shown in Fig.4a, which can be divided into three steps:

**3D Printing:** Four components are printed first: the main silicone body mold, the heating wire fixing mold, the magnets fixing mold, and the fixing frame for shear force. The *silicone body mold* has internal dimensions of approximately 20×20×15 mm and features a smooth transition at its base. The *heating wire mold* is designed with multiple columns to facilitate the wrapping of the heating wire. The *magnet fixing*

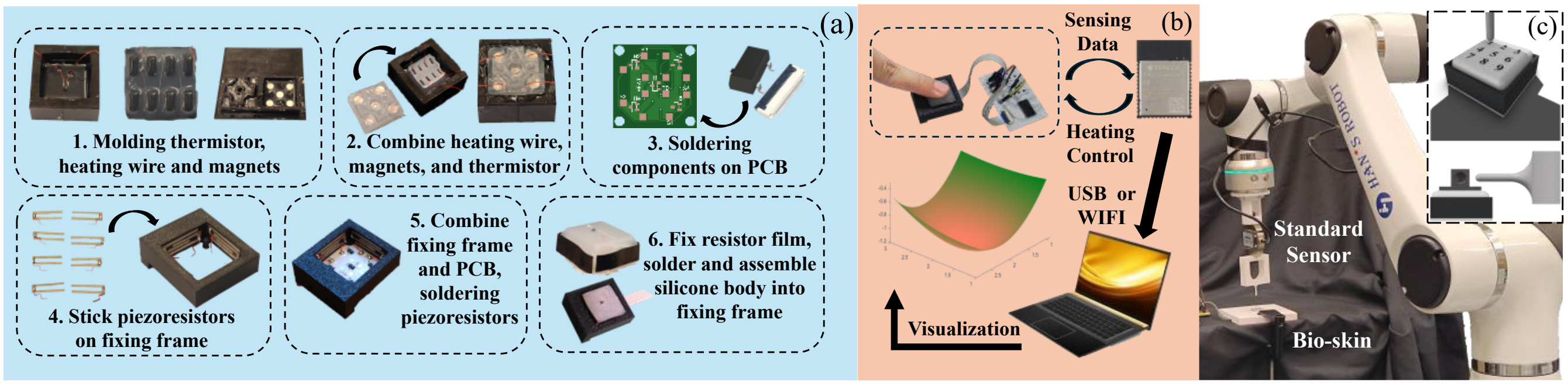

Fig.4 (a) Step-by-step fabrication process of Bio-Skin, including molding, assembly, and integration of key sensing components. (b) System architecture demonstrating real-time sensing, heating control, and data transmission via wired or wireless communication. (c) Calibration setup for collecting data of normal and shear force. Surface of Bio-Skin is divided into 1~9 positions and a direction converter is fixed on Bio-Skin to calibrate shear force.

*mold* is configured with two slots for soldering both sides of the magnets. One slot has four extra columns to leave the space for spring molding. The back of both slots is designed with five magnet slots which are used to attract and fix the magnets during molding processing.

**Molding: 1).** Mix the silicone gel in proper ratio. 2). Place the thermistor at the bottom of the mold and pour the gel to cover it fully. 3). Wrap the heating wire around its mold, leaving a 4 cm lead, and fill it with gel until it is covered. 4). Install magnets in two sets of slots on the back of the magnet-fixing mold, orienting them oppositely; then place them in the column slots and fill them with gel until fully encapsulated. 5). Allow the gel to solidify for 1 hour; then, invert the silicone gel with magnets into another slot, insert four springs into the column holes, and refill with gel to the mold's edge. 6). Once all parts are solidified, remove the heating wire and magnet silicone body from the mold, assemble the heating wire silicone body atop the thermistor silicone body, and position the magnet silicone body above them, filling any remaining voids with gel. 7). After 1 hour, remove the solidified silicone body, completing the molding process.

**Soldering and Assembly: 1).** Soldering resistors, capacitors, and Hall-effect sensors onto the customized sensor PCB. **2).** Fixing four film piezoresistors along the four sides of the fixing frame. **3).** Integrating the fixing frame with the sensor PCB and soldering the piezoresistors with the PCB. **4).** Soldering thermistor and heating wire to the PCB and embedding the silicone body into the fixing frame.

*F. I/O Configuration and Temperature Control*

**I/O:** A controller samples the Bio-Skin's analog signals and performs analog-to-digital conversion (ADC). Bio-Skin measures five normal force signals, four shear force signals, and one temperature signal (10 channels total). Another PCB board with a MUX chip is utilized to save ADC I/O resources. Bio-Skin connects to the MUX board via an FPC flex cable, and the board then connects to the controller. The controller switches MUX channels at 5 kHz, sampling 10 signal channels to form a complete data set at 500 Hz, which is transmitted to the host computer, as shown in Fig.4b.

**Temperature Control:** Temperature control is achieved through a feedback mechanism wherein the thermistor's reading modulates the heating power. Two threshold temperatures are defined: stop temperature ($T_s$) and heating temperature ($T_H$). When the thermistor detects that the temperature has reached $T_s$, the heating wire is deactivated; when the temperature falls below $T_H$, the heating wire remains continuously active. For temperature between $T_S$ and $T_H$, PWM modulation is implemented, with the duty cycle varying linearly from 0% to 100% to adjust the heating power.

## IV. CALIBRATION AND EVALUATION RESULTS

*A. Data Collection Method and Experiment Setup*

Calibration of Bio-Skin links raw voltage to force and temperature measurements. For force calibration, a six-axis robotic arm (Elfin E05, ±0.02 mm accuracy) applies forces to the Bio-Skin (Fig.4c) at various angles with a 3 mm tip hemispherical probe, ensuring precise force application. A 1-axis load cell (±1g sensitivity) is placed between the robot arm's end effector and the hemispherical probe to provide the ground truth force. The Bio-Skin and the load cell data are collected simultaneously for calibration. For temperature, semiconductor thermoelectric modules control the sensor's environment, while the thermistor signal is recorded and compared with thermal camera images for calibration.

For **normal force**, the sensor surface is divided into a 3×3 array, labeled No.1 – 9 shown in Fig.4c. The robotic arm applies a continuous force from 0 to 6 N at each point over five cycles, with each cycle collecting 12k samples from both the Bio-Skin and the standard force sensor. For **shear force**, a direction converter is mounted on the Bio-Skin to transfer shear force components, as shown in Fig.4c. The arm applies a continuous horizontal force from 0 to 10 N in one direction, then releases and shifts to the next until all four directions are covered, repeating for five cycles with 6k samples per cycle. For the **thermistor**, we turn off the heating function and use a semiconductor thermoelectric module to cool down and heat up the sensor with a 2V increase until 12V. Once the thermistor's output stabilizes, its reading is recorded, and a thermal camera (±0.1 °C accuracy) is used to measure the temperature as the ground truth, resulting in 13 samples total.

*B. Data Processing and Calibration*

**Data Processing:** Before calibration, raw force sensor data is pre-processed using a moving average (window size: 15) and a 4th-order Butterworth low-pass filter (sampling at 100Hz with a 5Hz cutoff), reducing signal fluctuation by ~8%, which is shown in Fig.5a. We partition three out of the five collected datasets as the calibration set, and two data set as the validation set to evaluate the accuracy calibrated sensor.

**Normal Force Sensor Calibration:** The calibrated normal force output will be the same as the data dimension, a 3X3 array to cover the Bio-Skin surface.

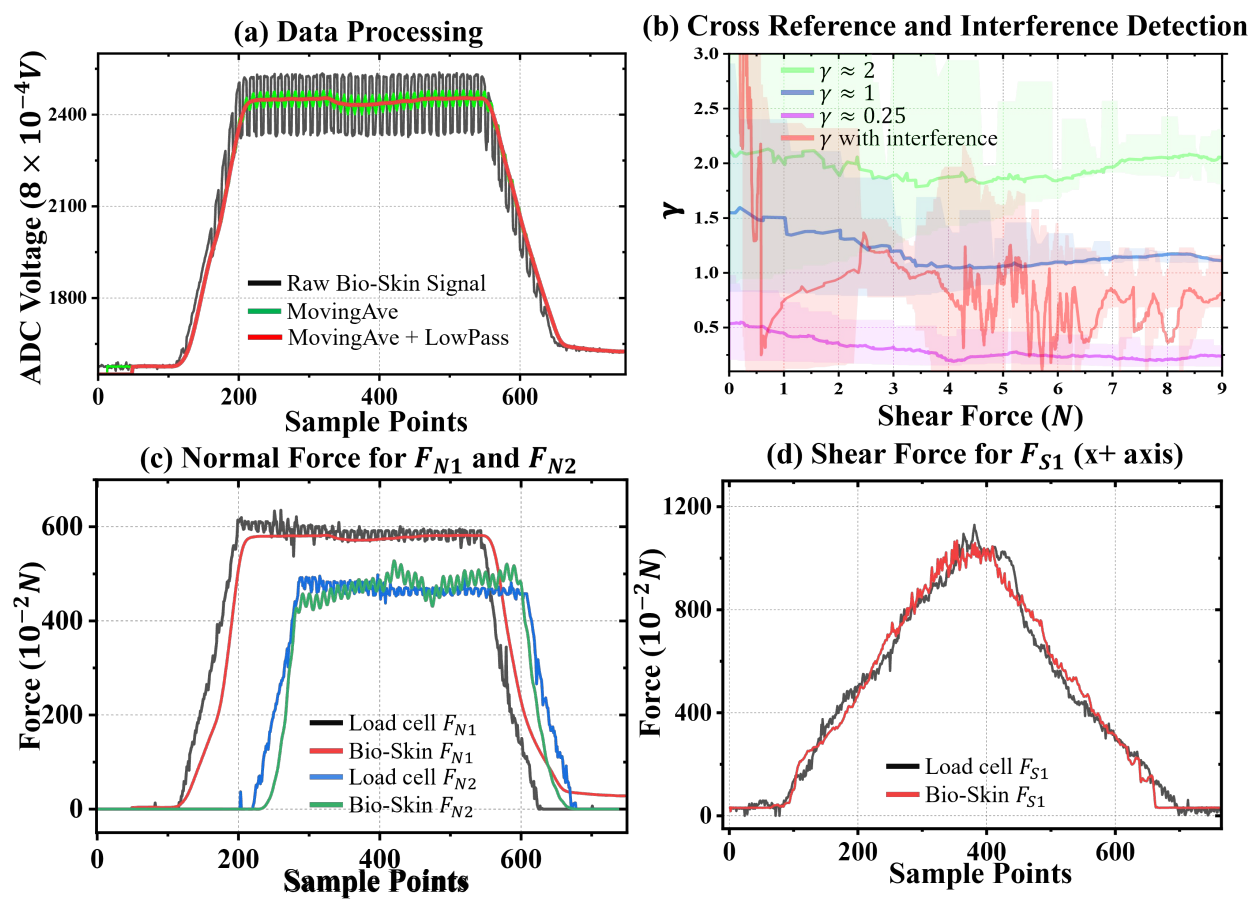

Fig.5. (a) compares the raw data of Bio-Skin and data after filtering. (b) shows the Bio-Skin's interference detection result, red line is detected under changed magnetic field and other 3 is detected under normal environment with different force direction. (c) compares the Bio-Skin's normal force and the ground truth at position 1 and 2, other positions have similar behaviors. (d) compares Bio-Skin's shear force and the ground truth in the positive x-axis, other directions have similar behaviors.

For the five normal force positions with Hall sensors (positions 1, 3, 5, 7, and 9) underneath, a fifth-order polynomial is defined to approximate the relation between the voltage output to the actual normal force:

$$F_{N_i} = b_i + \sum_{j=1}^{5} a_{ij} O_i^j \qquad i = 1,3,5,7,9 \qquad (1)$$

where $F_{N_i}$ is the normal force for Hall-effect sensor $i$, $b_i$ is the bias, $a_{ij}$ is the parameter for the fifth-order polynomial, $O_i^j$ represents the filtered output of position $i$.

For positions (2, 4, 6, and 8), normal force is calibrated using a multivariate linear calibration based on the fitted values from the three adjacent positions. For instance, region 2 will be calibrated using the values from regions 1, 3, and 5. The generalized representation is as follows:

$$F_{N_i} = b_i + \sum_{j=1}^{3} d_{ij} \cdot F_{r_j} \qquad i = 2,4,6,8 \qquad (2)$$

where $d_{ij}$ represents the coefficients of the multivariate linear function, $F_{r_j}$ represents three adjacent forces surrounding the position of (2, 4, 6, 8) in a 3X3 configuration.

**Shear Force Sensor Calibration:** a fourth-order polynomial is defined to approximate the relation between the resistance output to the actual shear force:

$$F_{S_k} = x_k + \sum_{j=1}^{4} y_{ij} O_k^j \qquad k = 1,2,3,4 \qquad (3)$$

where $F_{S_k}$ is the shear force for each side of the Bio-Skin shell, $x_k$ is the bias, $y_{ij}$ is the parameter for the fourth-order polynomial. $O_k^j$ represents the filtered output of direction $k$, in the order of x+, x-, y+, y- direction.

**Temperature Sensor Calibration:** From the collected 13 samples of thermistor and thermal camera output $\tau_n$, where $n = 1,2, \dots 12$, we can derive 12 piecewise linear functions to approximate the nonlinear relationship between the thermistor signal and the temperature change. The functions are defined as:

$$T = \alpha_n R + \beta_n \qquad (4)$$

where $T$ is the temperature in Celsius, $R$ represents the thermal resistance signal sampled by the controller, $\alpha_n$ and $\beta_n$ are the piecewise linear function coefficients, which are selected for each piece based on the threshold $\tau_n < T < \tau_{n+1}$.

TABLE II. BIO-SKIN PERFORMANCE AFTER CALIBRATION

| | RMSE (N) | MAE (N) | Noise to Range Ratio | $R^2$ |
|---|---|---|---|---|
| Normal (1,3,5,7,9) | 0.26 | 0.19 | ≈ 3.3% | 0.991 |
| Normal (2,4,6,8) | 0.42 | 0.23 | ≈ 4.7% | 0.963 |
| Shear | 0.47 | 0.37 | ≈ 3.7% | 0.981 |

TABLE III. TECHNICAL SPECIFICATION COMPARISON WITH USKIN

| | Z-axis Range (N) | XY-axis Range (N) | Sampling Frequency (Hz) | Interference Detection | Noise to Range Ratio | Estimated Cost (USD) |
|---|---|---|---|---|---|---|
| Bio-Skin | 6 (single point) >12 (surface) | ±10 | 500 | Yes | ≈ 3.8% | <2000/each |
| uSkin | 7.3 | ±0.7 | 71 | No | ≈ 0.12% | 70000/Hand |

**Validation Results:** Bio-Skin's performance is evaluated in the testing set after calibration. The plots are shown in Fig.5c and Fig.5d. The statistical results are shown in Table. II. Bio-Skin demonstrates excellent accuracy in both normal and shear force, with RMSE as low as 0.26N, maximum noise to range ratio is lower than 4.7% and coefficient of determination ($R^2$) of all approximations over 0.96. The built-in thermistor matches the data from the thermal camera and defaults to 33°C. Bio-Skin's technical specifications are compared with the uSCu sensor from uSkin in Table III. Bio-Skin shows significant advantages in sampling frequency, shear force detection range, interference detection, and cost.

### *C. Cross Reference Validation*

Bio-Skin's multi-modal design allows different sensing modalities to cross-reference each other to validate signal fidelity, which is a critical function in complex electromagnetic environments[16]. Thus, in this section, we develop a cross-reference framework between the Hall-effect sensor and piezoresistors in force measurement. According to Newton's law, the normal and shear forces on Bio-Skin are fractions of the direct force. Thus, we can assume a certain relationship exists between these two forces, even with friction loss. Such a relationship is directly correlated to the sensor signal. We define a cross-reference coefficient $\gamma$ by taking the fraction of the calibrated normal and shear force to model a linear relationship between these two forces:

$$\gamma = F_N / F_S \qquad (5)$$

To validate the Cross Reference model, we collect data using a flat plate with a central depression that was placed on the Bio-Skin surface to facilitate uniform force application. Then, the robotic arm applies a certain force on the Bio-Skin surface at three different force directions, so the $\gamma$ ranges from 0.25 to 2, resulting in a total of 30k valid samples. Then, we apply a changing magnetic field around the Bio-Skin to simulate the interference. The result of $\gamma$ under different force angles and environments is shown in Fig.5b.

Without interference, $\gamma$ achieves a relatively stable value, especially when the shear force is over 4N. However, in a magnetically changing environment, $\gamma$ fails to converge. The results prove that the multi-modality of Bio-Skin can be used to detect electromagnetic interference and verify the accuracy of the Hall sensor data.

### *D. GUI and Visualization*

To improve the usability and visualize the data from Bio-Skin, we designed a graphical interface and control panel for calibrating and monitoring. The system includes: 1). a GUI that displays the current force data from the sensor, allows for independent zero calibration of both normal and shear force, and records the sensor's force and temperature data; and 2). a visualization that simulates the force distribution on the Bio-Skin surface through modeling, displaying the actual force conditions in real-time, which is shown in Fig.6.

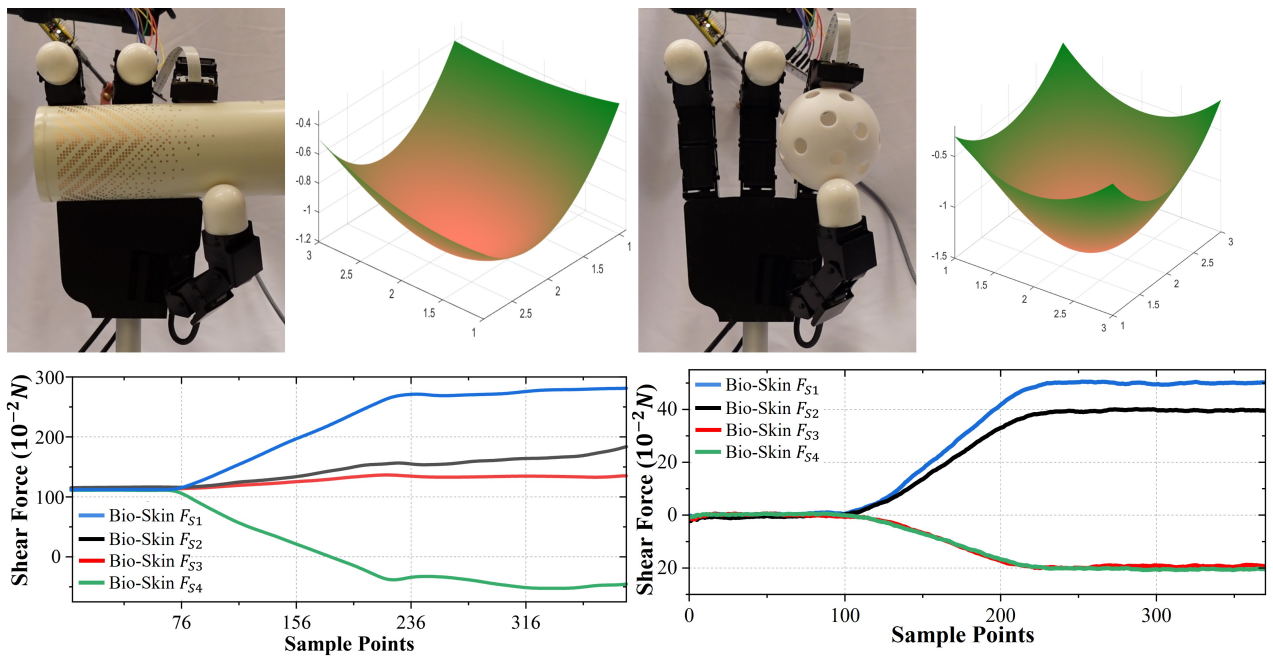

Fig.6. The force visualization of Allegro Hand fingertip with Bio-Skin while grasping a plastic ball (right) and a metal cup (left), showing distinct normal force distribution and shear force changing.

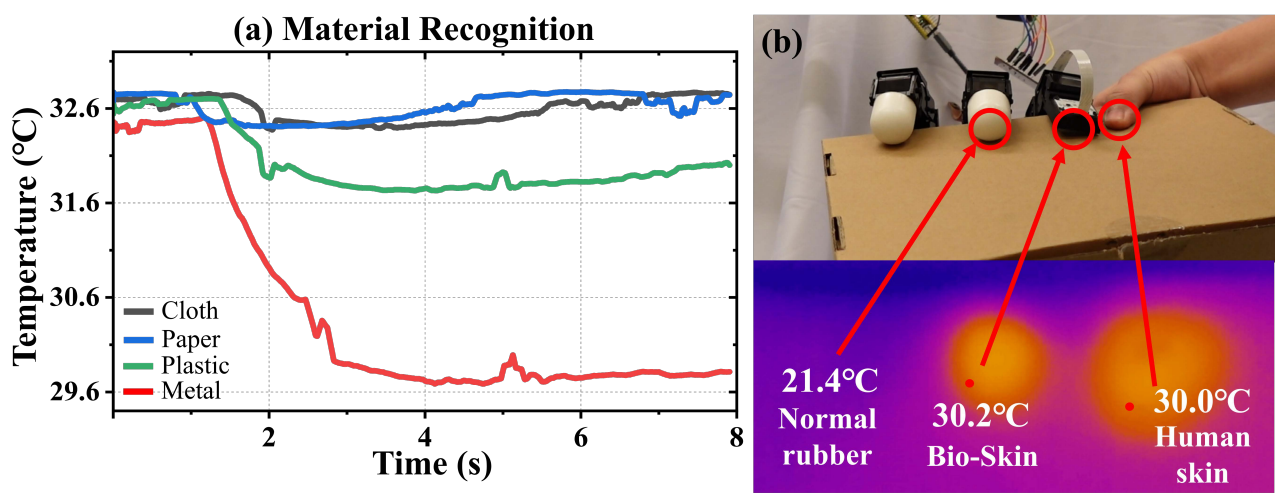

Fig.7. (a) shows the temperature changing after Bio-Skin touch different materials with temperature regulation on. (b) shows temperature retention by comparing Bio-Skin, human skin and normal rubber.

### *E. Material Recognition*

Leveraging its temperature sensing, Bio-Skin can detect the temperature dissipation rate upon contact, offering an extra modality for material identification due to varying thermal conductivities. To evaluate this, Bio-Skin stabilized at 32°C and then touched the plastic, cardboard, fiber cloth, and metal at 26°C environment temperature. The result is shown in Fig.7a. When Bio-Skin contacts metal, its temperature drops rapidly to 29°C and recovers very slowly. With plastic, a rapid drop also occurs but stabilizes at about 31.6°C, with recovery starting after 6 seconds. In contrast, contact with cardboard and fiber cloth causes only a 0.4°C drop, with recovery beginning within 2 seconds. The temperature measuring function of Bio-Skin thus further enables material identification in an extra modality at the same initial temperature. Compared to the current no-heating solution [17], the heating function of Bio-Skin produces a larger temperature gap between the sensor and environment, resulting in more sensitive measuring of temperature change.

### *F. Grasping Test*

To verify the potential for further development and practical application of Bio-Skin, we installed it on an Allegro Hand, which was controlled to grasp a plastic ball and a metal cup while observing the visualization interface (shown in Fig.6). The results show that when grasping the plastic ball, the force was concentrated in the central area. In contrast, the force distribution was more uniform when grasping the cup. The shear force also varies due to the gripping angle. The results prove that the 3D planner force detection and shear force detection can be used to identify different contact surfaces and, hence, different shapes of objects.

### *G. Contact Temperature Retention*

To demonstrate another potential benefit of Bio-Skin's temperature regulation function, we controlled the Allegro Hand with Bio-Skin grasp a piece of cardboard for 10 seconds while a human subject simultaneously did the same task. Thermal imaging was then used to observe the retained temperature on the cardboard, as shown in Fig.7b. While differences in contact area led to variations in the temperature retention range, the core temperatures were very similar, proving Bio-Skin's ability to mimic the thermal print of human skin in temperature retention.

## REFERENCES

[1]. J. A. Fishel and G. E. Loeb, "Sensing tactile microvibrations with the BioTac—Comparison with human sensitivity," in 2012 4th IEEE RAS & EMBS international conference on biomedical robotics and biomechatronics (BioRob), IEEE, 2012, pp. 1122–1127.

[2]. T. P. Tomo *et al.*, "Covering a robot fingertip with uSkin: A soft electronic skin with distributed 3-axis force sensitive elements for robot hands," *IEEE Robot. Autom. Lett.*, vol. 3, no. 1, pp. 124–131, 2017.

[3]. T. P. Tomo *et al.*, "A new silicone structure for uSkin—A soft, distributed, digital 3-axis skin sensor and its integration on the humanoid robot iCub," *IEEE Robot. Autom. Lett.*, vol. 3, no. 3, pp. 2584–2591, 2018.

[4]. J. Di *et al.*, "Using fiber optic bundles to miniaturize vision-based tactile sensors," *IEEE Trans. Robot.*, 2024.

[5]. R. W. Van Boven and K. O. Johnson, "The limit of tactile spatial resolution in humans: grating orientation discrimination at the lip, tongue, and finger," Neurology, vol. 44, no. 12, p. 2361, 1994.

[6]. H. S. Tsien, "Engineering cybernetics," 1954.

[7]. X. Li et al., "A high-sensitivity magnetic tactile sensor with a structure-optimized Hall sensor and a flexible magnetic film," IEEE Sens. J., 2024.

[8]. T. P. Tomo et al., "Design and characterization of a three-axis hall effect-based soft skin sensor," Sensors, vol. 16, no. 4, p. 491, 2016.

[9]. P. Mittendorfer, E. Yoshida, and G. Cheng, "Realizing whole-body tactile interactions with a self-organizing, multi-modal artificial skin on a humanoid robot," Adv. Robot., vol. 29, no. 1, pp. 51–67, 2015.

[10]. B. Romero, F. Veiga, and E. Adelson, "Soft, round, high resolution tactile fingertip sensors for dexterous robotic manipulation," in 2020 IEEE International Conference on Robotics and Automation (ICRA), IEEE, 2020, pp. 4796–4802.

[11]. S. Zhang et al., "Hardware Technology of Vision-Based Tactile Sensor: A Review," in IEEE Sensors Journal, vol. 22, no. 22, pp. 21410-21427, 15 Nov.15, 2022.

[12]. S. Gao, L. Weng, Z. Deng, B. Wang, and W. Huang, "Biomimetic tactile sensor array based on magnetostrictive materials," *IEEE Sens. J.*, vol. 21, no. 12, pp. 13116–13124, 2021.

[13]. R. Bhirangi, T. Hellebrekers, C. Majidi, and A. Gupta, "Reskin: versatile, replaceable, lasting tactile skins," in *5th Annual Conference on Robot Learning*, 2021.

[14]. P. Weiner, C. Neef, Y. Shibata, Y. Nakamura, and T. Asfour, "An embedded, multi-modal sensor system for scalable robotic and prosthetic hand fingers," Sensors, vol. 20, no. 1, p. 101, 2019.

[15]. P. Mittendorfer and G. Cheng, "Humanoid multi-modal tactile-sensing modules," *IEEE Trans. Robot.*, vol. 27, no. 3, pp. 401–410, 2011.

[16]. S. B. Barnett, A. G. Swanson, T. Lorimer, and M. Brown, "Electromagnetic interference mitigation in a high voltage inspection robot," in Proceedings of the 21st International Symposium on High Voltage Engineering: Volume 1, Springer, 2020, pp. 331–341.

[17]. A. G. Eguíluz, I. Rañó, S. A. Coleman, and T. M. McGinnity, "A multi-modal approach to continuous material identification through tactile sensing," in *2016 IEEE/RSJ International Conference on Intelligent Robots and Systems (IROS)*, IEEE, 2016, pp. 4912–4917